\documentclass{article}
\usepackage[numbers]{natbib}
\PassOptionsToPackage{numbers, compress}{natbib}

\usepackage[final]{nips_2018}

\usepackage[utf8]{inputenc} % allow utf-8 input
\usepackage[T1]{fontenc}    % use 8-bit T1 fonts
\usepackage{hyperref}       % hyperlinks
\usepackage{url}            % simple URL typesetting
\usepackage{booktabs}       % professional-quality tables
\usepackage{amsfonts}       % blackboard math symbols
\usepackage{nicefrac}       % compact symbols for 1/2, etc.
\usepackage{microtype}      % microtypography
\usepackage{booktabs}
\usepackage{graphicx}
\usepackage{subfig}
\usepackage{mathtools}  
\usepackage{color}
\usepackage{multirow}

\title{Generative Models for Simulating Mobility Trajectories}

\author{
  Vaibhav Kulkarni\\
\ \  Department of Information Systems\\ 
  UNIL-HEC Lausanne\\
  \texttt{vaibhav.kulkarni@unil.ch} \\
  \And
  Natasa Tagasovska \\
\ \  Department of Information Systems\\ 
  UNIL-HEC Lausanne\\
  \texttt{natasa.tagasovska@unil.ch}\\
  \AND
  Thibault Vatter \\
\ \ \ \ \ \ \ \ \ Department of Statistics \ \ \ \ \ \ \ \ \  \ \\  
Columbia University\\
  \texttt{tv2233@columbia.edu} \\
  \And
  Benoit Garbinato \\
 \ \  Department of Information Systems\\
  UNIL-HEC Lausanne \\
  \texttt{benoit.garbinato@unil.ch} \\
}

\begin{document}

\maketitle

\begin{abstract}

Mobility datasets are fundamental for evaluating algorithms pertaining to geographic information systems and facilitating experimental reproducibility.
But privacy implications restrict sharing such datasets, as even aggregated location-data is vulnerable to membership inference attacks. 
Current synthetic mobility dataset generators attempt to superficially match {\it{a priori}} modeled mobility characteristics which do not accurately reflect the real-world characteristics.
Modeling human mobility to generate synthetic yet \emph{semantically and statistically realistic} trajectories is therefore crucial for publishing trajectory datasets having satisfactory utility level \emph{while preserving user privacy}. 
Specifically, long-range dependencies inherent to human mobility are challenging to capture with both discriminative and generative models.  
In this paper, we benchmark the performance of recurrent neural architectures (RNNs), generative adversarial networks (GANs) and nonparametric copulas to generate synthetic mobility traces. %this is fairly obvious trained on a real-world mobility dataset. 
We evaluate the generated trajectories with respect to their geographic and semantic similarity, circadian rhythms, long-range dependencies, training and generation time.
We also include two sample tests to assess statistical similarity between the observed and simulated distributions, and we analyze the privacy tradeoffs with respect to membership inference and location-sequence attacks.

\end{abstract}

\section{Introduction}

The pervasiveness of mobile devices equipped with internet connectivity and global-positioning functionality has resulted in an increasingly large amount of location-data on individuals.
This data is beneficial to address and validate spatiotemporal data-based problems; predictive and kNN queries, object tracking, mobility modeling and location privacy among others.
Due to the sensitive nature of datasets containing mobility traces, sharing them with untrusted entities present privacy implications.
Trivial heuristics can be applied on such datasets to derive personally identifiable information of individuals, even at aggregate levels~\citep{xu2017trajectory}.

Publicly accessible mobility datasets~\citep{zheng2010geolife, mokhtar2017priva, Nokiadataset1} are usually not adequate for large scale experimental evaluations, compromising scalability tests.
This issue incentivizes synthetic mobility trajectory generators that simulate the behavior of moving objects required to attain comprehensive performance valuations. 
In this context, one typically considers rigid and unnatural mobility models, not guaranteeing the existence or even cardinality of patterns within the synthetic population.
Alternative approaches rely on parametric sequential models~\citep{Kulkarni:2017:GSM:3149808.3149809} and Markov processes~\citep{bindschaedler2016synthesizing} to learn and generate trajectories.
Such techniques also ignore the presence of long-range dependencies\cite{lin2016critical} inherent to human mobility which features non-Markovian character~\citep{zhao2015non, kulkarni2018inability}.

It is therefore imperative to generate context-dependent synthetic traces resembling the human-mobility behavior at satisfactory utility levels while preserving user privacy. 
However, one of the major challenges is the absence of quantitative methods for evaluating the realistic nature of synthetic traces and the associated utility-privacy tradeoff.

To this end, we present several nonparametric approaches to generate large-scale synthetic trajectories by training the models on a real-world dataset followed by hallucinating trajectories using the trained model.
We perform an extensive evaluation of the generated trajectories by assessing their geographic and semantic similarity compared to the actual dataset.
We use two sample metrics to obtain the statistical similarity between datasets. 
We then quantify the presence of long-range dependencies by computing the mutual-information decay and conduct privacy-leakage tests on the generated trajectories.  
We conclude with a discussion on appropriate strategies and applicable evaluation metrics based on our experimental results and tackle open questions and challenges.

\section{Related Work}

\autoref{tbl:related_work_summary} provides a summary of existing trajectory generators, where they formulate the synthetic trajectory simulation as an optimization problem, solved by genetic algorithms under the constraint of {\it{a priori}} determined parameters.
A fundamental issue is the selection and definition of the parameter space that controls the evolution of the moving objects.
The stringent and classified network connections thus influence the realistic nature of the generated trajectories.
In several cases, there is no correlation between the future direction of movement and the past locations.
Repeated visits to a given location within a short span of time are also observed due to the bounding parameters. 
Therefore, the symbolic nature of these frameworks result in an implicit location-dependent context, which compromises the realistic nature of the generated activity patterns.  
To address these drawbacks associated with parametric modeling,~\citet{ouyang2018non} propose a GAN-based approach to generate trajectories, where the discriminator is based on a convolutional neural network (CNN)~\cite{lecun1995convolutional}.
Similarly, we explore other deep learning architectures based on RNNs known to model sequential data better than CNNs~\cite{schmidhuber2015deep}.
We also investigate generative models based on the nonparametric copulas of \citep{Geenens2017}.

\begin{table}[t!]
\centering
\caption{Categorization of current approaches to generate synthetic trajectories and parameters.}
\resizebox{\textwidth}{!}{%
\begin{tabular}{lll}
\hline
\textbf{Technique} & \textbf{Model name} & \textbf{Parameters considered} \\ \hline
\multirow{3}{*}{\textbf{Free movement}} & GSTD~\cite{theodoridis1999generation} & statistical distributions (mean, skew, standard deviation) \\
 & G-TERD~\cite{nascimento2003synthetic} & speed, rotation-angle, direction \\
 & Oporto~\cite{giannotti2005synthetic} & start time, end time, velocity, orientation \\ \hline
\multirow{5}{*}{\textbf{Road networks}} & Brinkoff~\cite{brinkhoff2002framework} & speed, street capacity, nearby object location, shortest path \\
 & SUMO~\cite{behrisch2011sumo} & road length, headway time, lane change times \\
 & BerlinMOD~\cite{duntgen2009berlinmod} & road network, trip start and end, Brinkoff model \\
 & ST-ACTS~\cite{gidofalvi2006st} & Geo-dependency model \\
 & Hermoupolis~\cite{pelekis2015hermoupolis} & mobility pattern, road network, points of interest \\ \hline
\multirow{2}{*}{\textbf{Multi environments}} & MWGen~\cite{xu2012mwgen} & trip plan, road network, floor plan, routing graph \\
 & MNTG~\cite{mokbel2008sole} & movement model, moving objects, simulation time \\ \hline
\multirow{2}{*}{\textbf{\begin{tabular}[c]{@{}l@{}}Sequential models\end{tabular}}} & Markov models~\cite{bindschaedler2016synthesizing} & semantic locations, geographic \\
 & Semi-Markov models~\cite{baratchi2014hierarchical} & stay points, transition paths \\ \hline
\end{tabular}%
}
\label{tbl:related_work_summary}
\end{table}

\section{Synthesizing Trajectories using Generative Modeling}

First we explore the benefits of applying deep learning architectures to synthesize mobility trajectories.
RNNs use their hidden memory representation to process input sequences and we select four architectures: (1) Char-RNN (SRNN)~\citep{grossberg2013recurrent}, (2) RNN-LSTM~\citep{Hochreiter1997LongSM}, (3) recurrent highway networks (RHN)~\citep{Zilly2017RecurrentHN}, and (4) pointer sentinel mixture model (PSMM)~\citep{Merity2016PointerSM}.
For GANs, where two neural networks compete in a zero-sum game framework, we select two architectures: (1) SGAN~\citep{yu2017seqgan}, and (2) RGAN~\citep{esteban2017real}.
These architectures differ in their capacity to manipulate their internal memory representation and propagate gradients along the network.
In addition to neural-network based solutions, we also evaluate \emph{copulas}; a seldom explored generative model in the machine learning community.
Given a bivariate random vector $(X_1, X_2)$, Sklar's theorem~\citep{sklar1959} states that the joint density\footnote{It is usually stated for the distribution rather than the density and for random vectors of arbitrary dimension.} is
$f(x_1, x_2) =  f_1(x_1)  f_2(x_2) c(F_1(x_1), F_2(x_2))$, where $f$ and $f_i$ are the marginal densities, $F_i$ the marginal distributions, and $c$ the copula density.
In other words, the bivariate density can be uniquely described by the product between its marginal densities and a copula density representing its dependence structure.
A useful consequence of this representation is that, by taking the logarithm on both sides, estimation of the joint density can be performed in two steps: the marginal distributions first, and the copula afterwards.
In a nutshell, copulas allow to flexibly specify the marginal and joint behavior of random variables.

An important aspect in the generative context is that, because $U=F(X) \sim U(0,1)$ for any continuous random variable with distribution $F$,  the copula is a distribution with uniform margins.
Hence, from a copula sample $(U_1, U_2)$, one obtains a sample on the original scale using the inverse cumulative distributions via $(X_1, X_2) = (F_1^{-1}(U_1), F_2^{-1}(U_2))$.
For further details on copulas, we refer the reader to~\citep{Joe14}.
In this paper, we combine the kernel-based nonparametric copulas of~\citep{Geenens2017} with the empirical distribution function of the margins obtain highly flexible models.

\textbf{Data representation} Given a dataset of $n$ mobility trajectories, where a trajectory $T_u$ of an individual $u$ is a temporally ordered sequence of tuples, such that, $T_u = \langle (l_1,t_1),(l_2,t_2)...(l_n,t_n)\rangle$, where $l_i = (lat_i,lon_i), 0 \leq i \leq n$, the latitude-longitude coordinate pair and $t$, the timestamp such that $t_{i+1} > t_{i}$.
We first transform the location data onto a uniform grid for dimensionality reduction using a technique that preserves spatial locality\footnote{Google S2: https://s2geometry.io/}, thus translating $T_u$ into a 2-D trace $S(t) = \langle (c_1,t_1),(c_2,t_2)...(c_n,t_n)\rangle$, where $c_i$ is the {\it{geo-hash}} of the projected cell ID and the timestamp $t_i$.

\section{Experiments, Results and Discussion}

\begin{figure}[!h]
\resizebox{\textwidth}{!}{
    \centering
    \subfloat[Char-RNN]{{\includegraphics[scale=0.1]{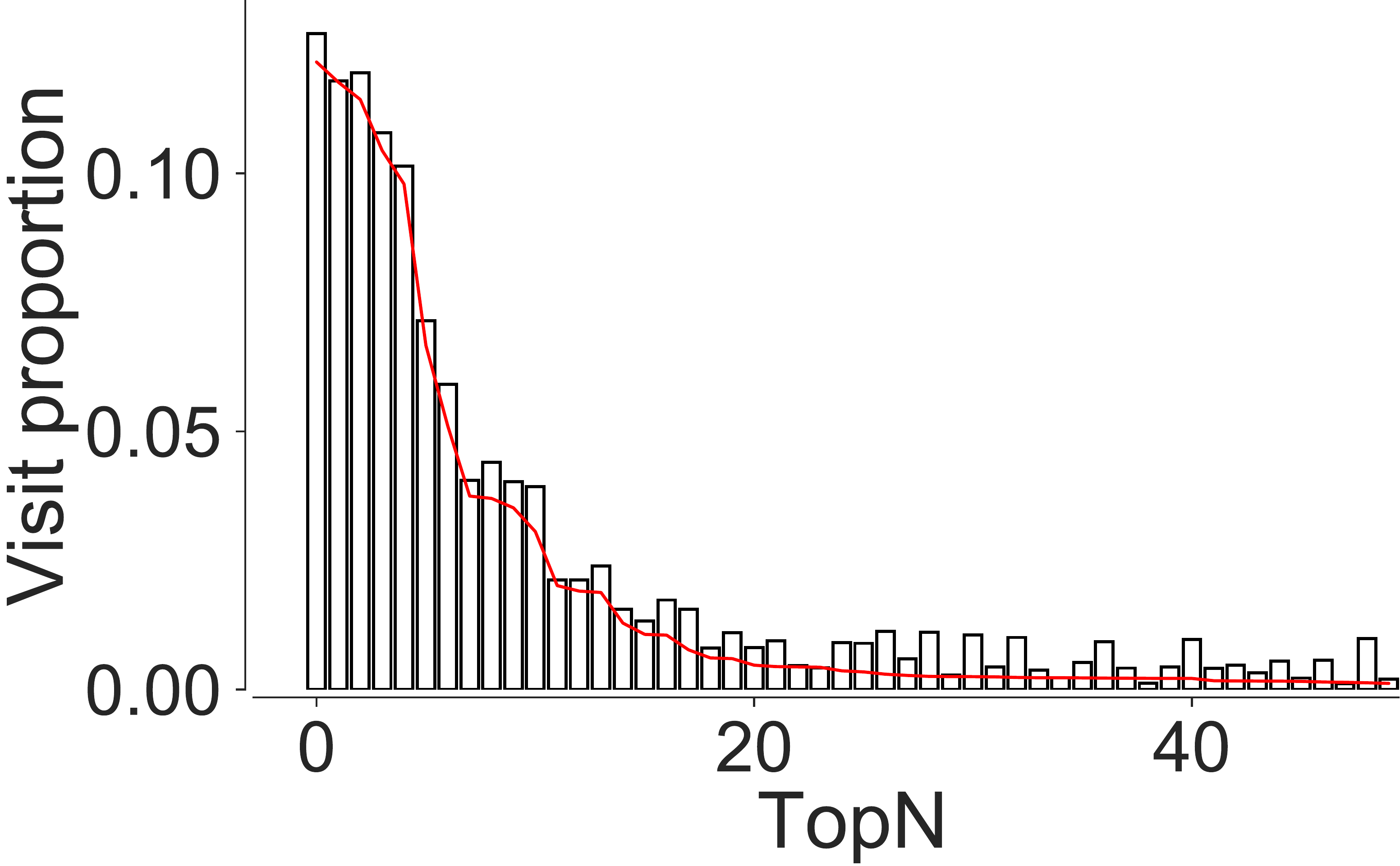} }}%
    \subfloat[RNN-LSTM]{{\includegraphics[scale=0.1]{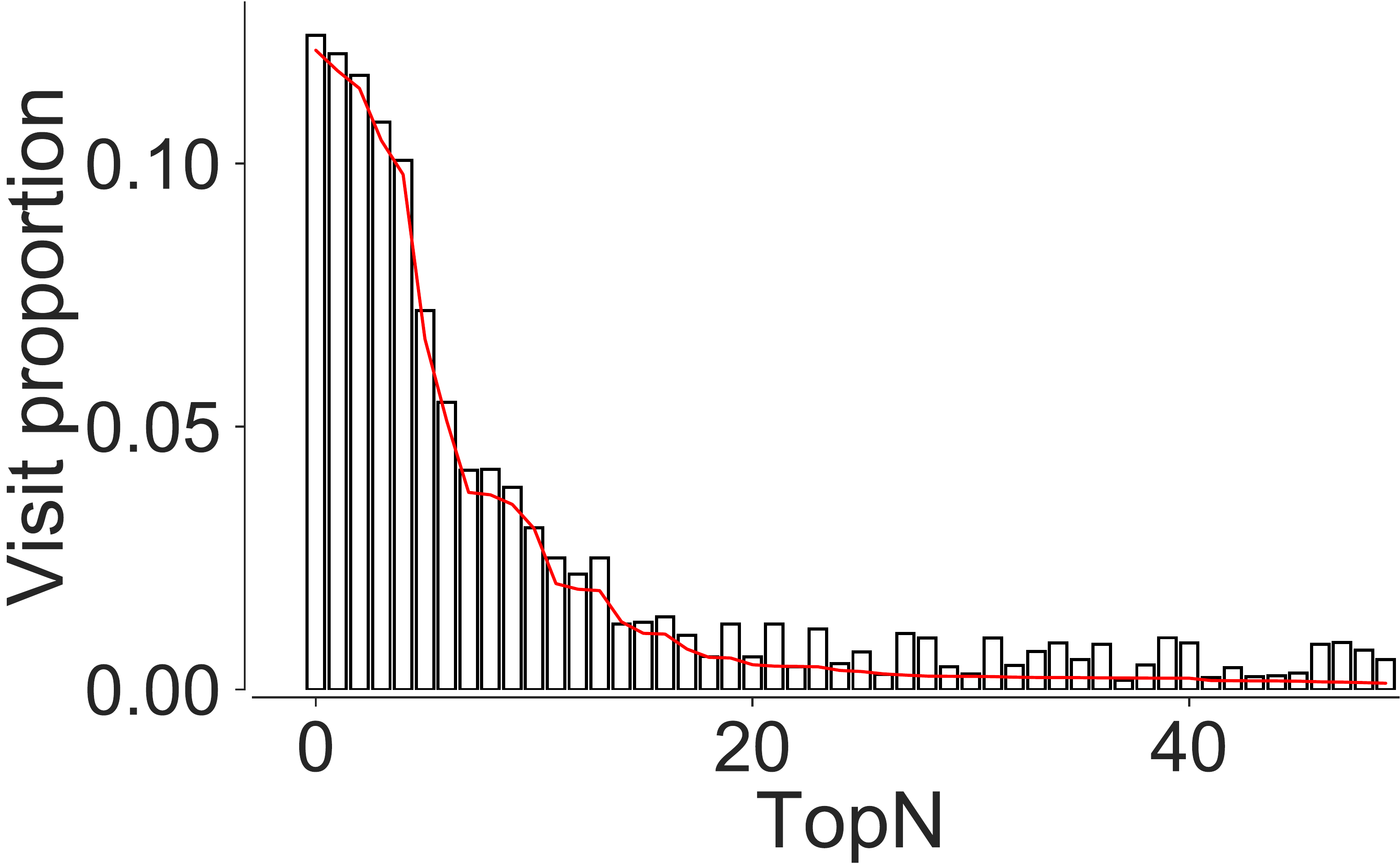} }}%
    \subfloat[RHN]{{\includegraphics[scale=0.1]{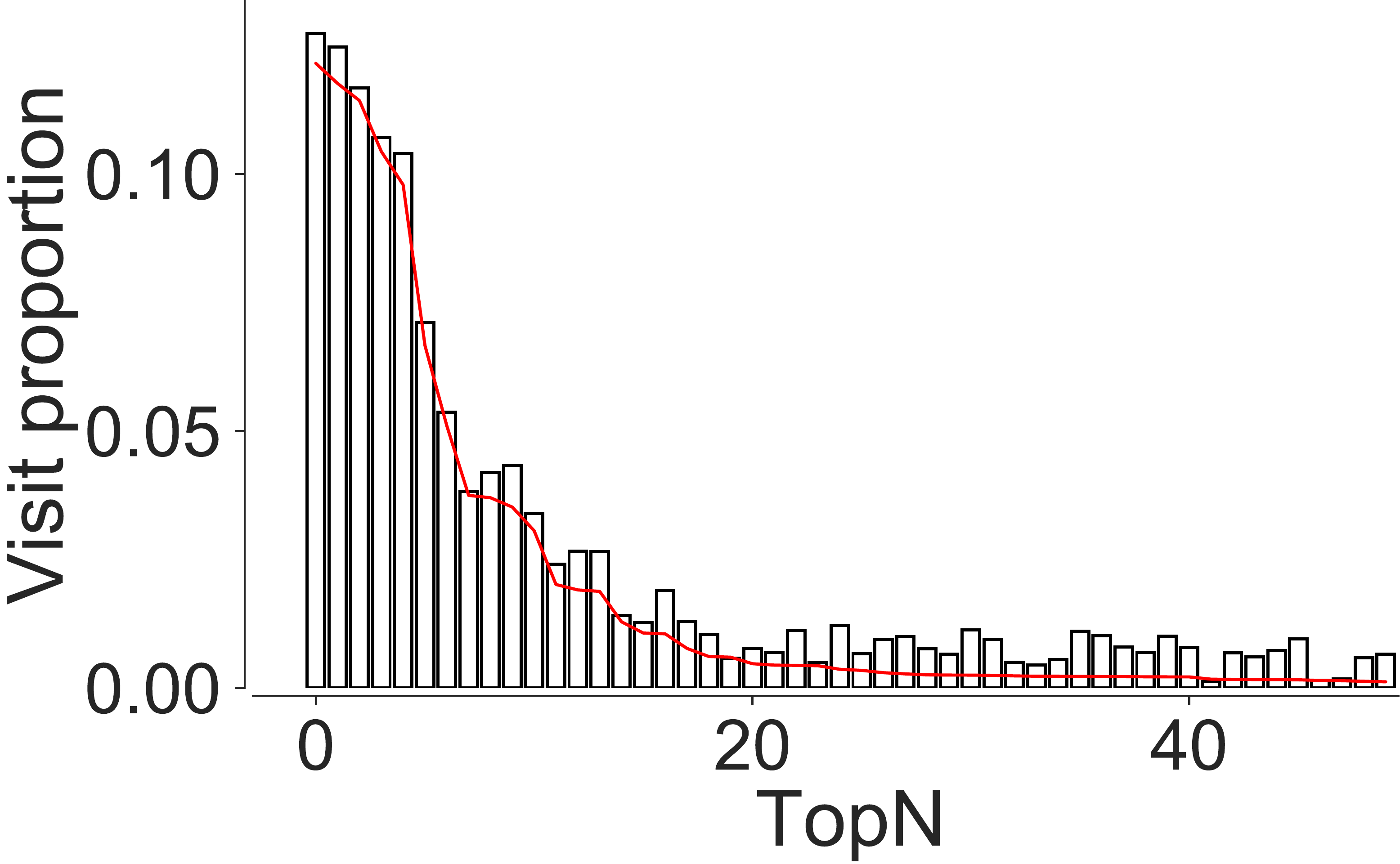} }}\newline
    \subfloat[PSMM]{{\includegraphics[scale=0.1]{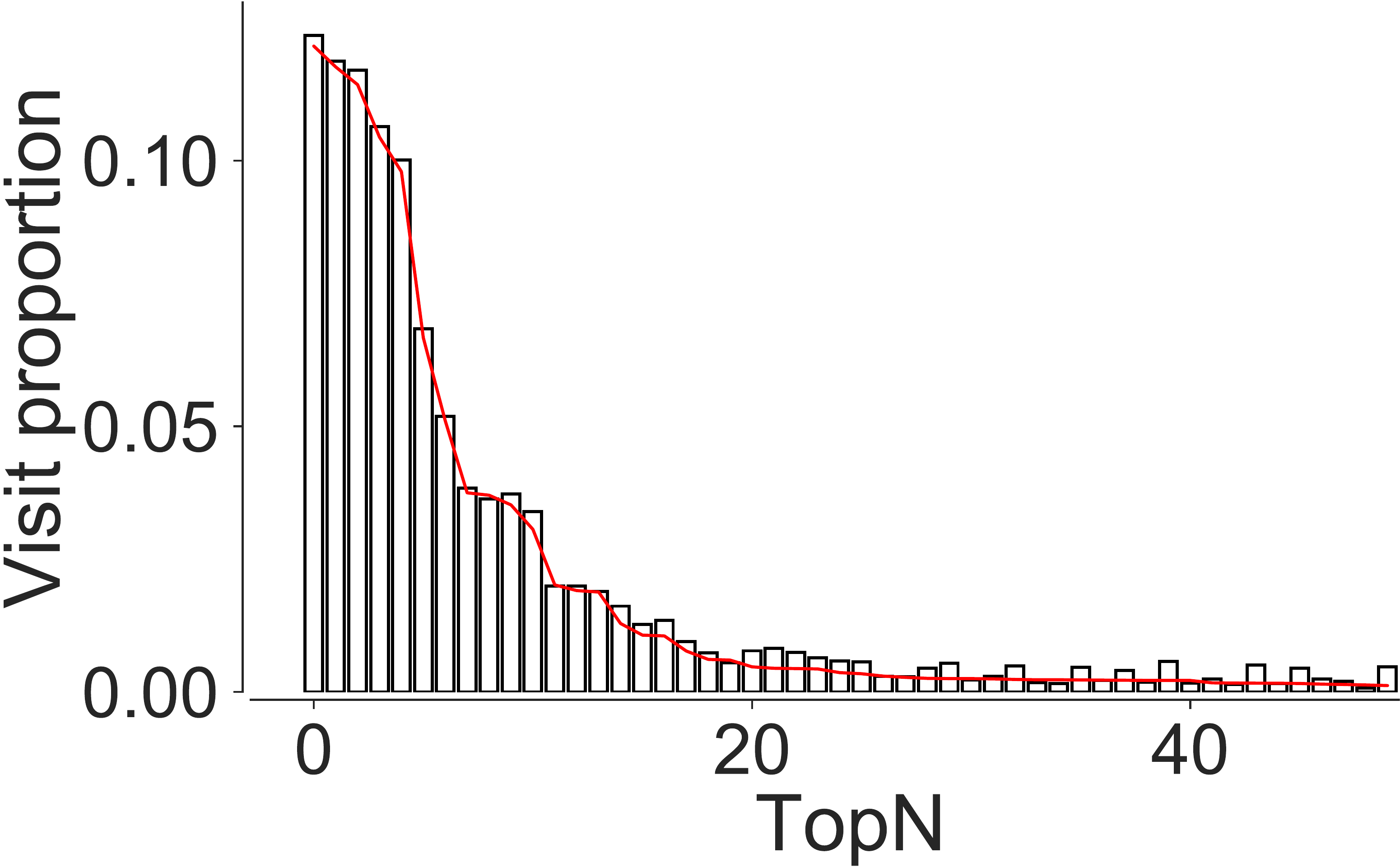} }}
    \subfloat[SGAN]{{\includegraphics[scale=0.1]{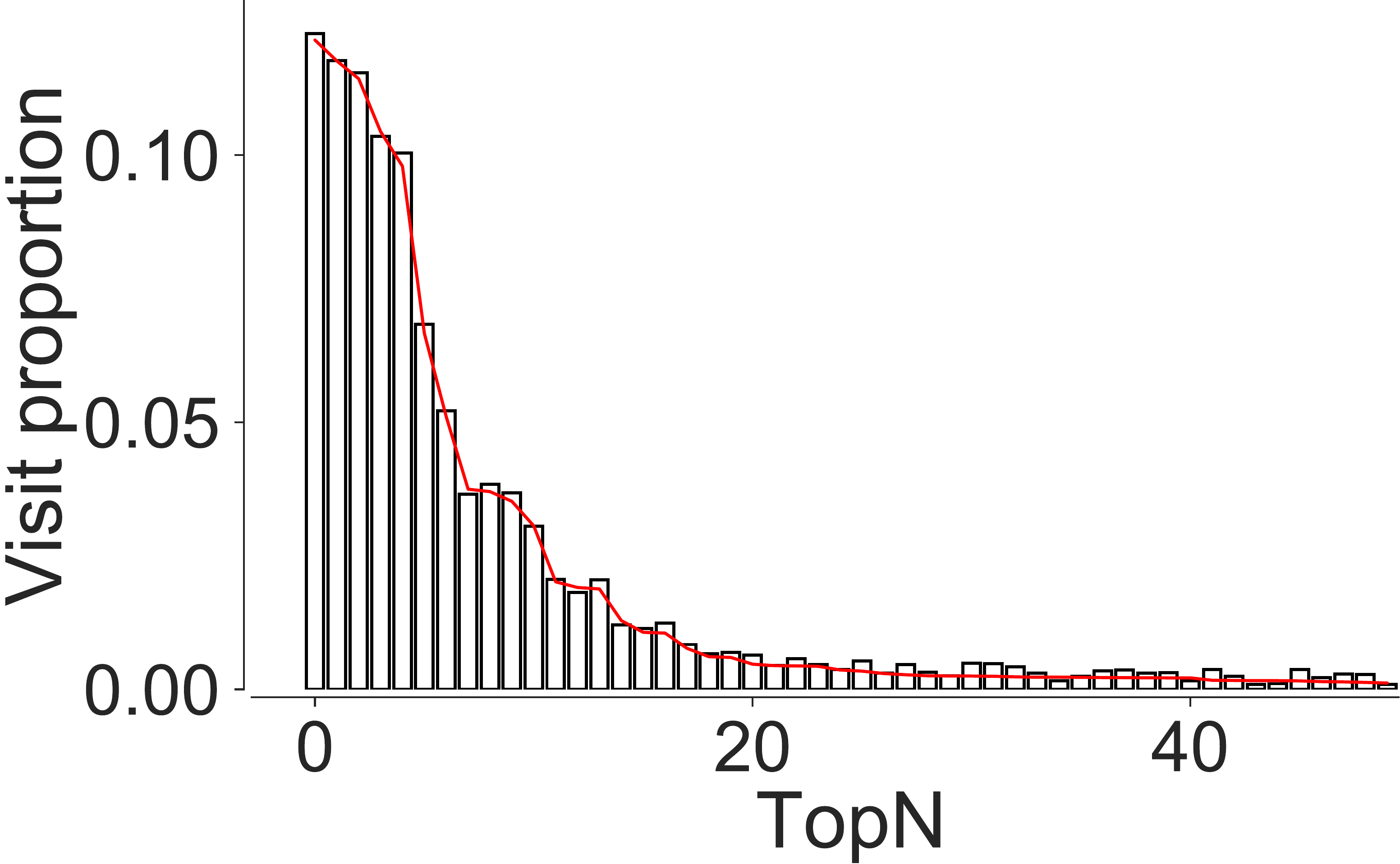} }}%
    \subfloat[RGAN]{{\includegraphics[scale=0.1]{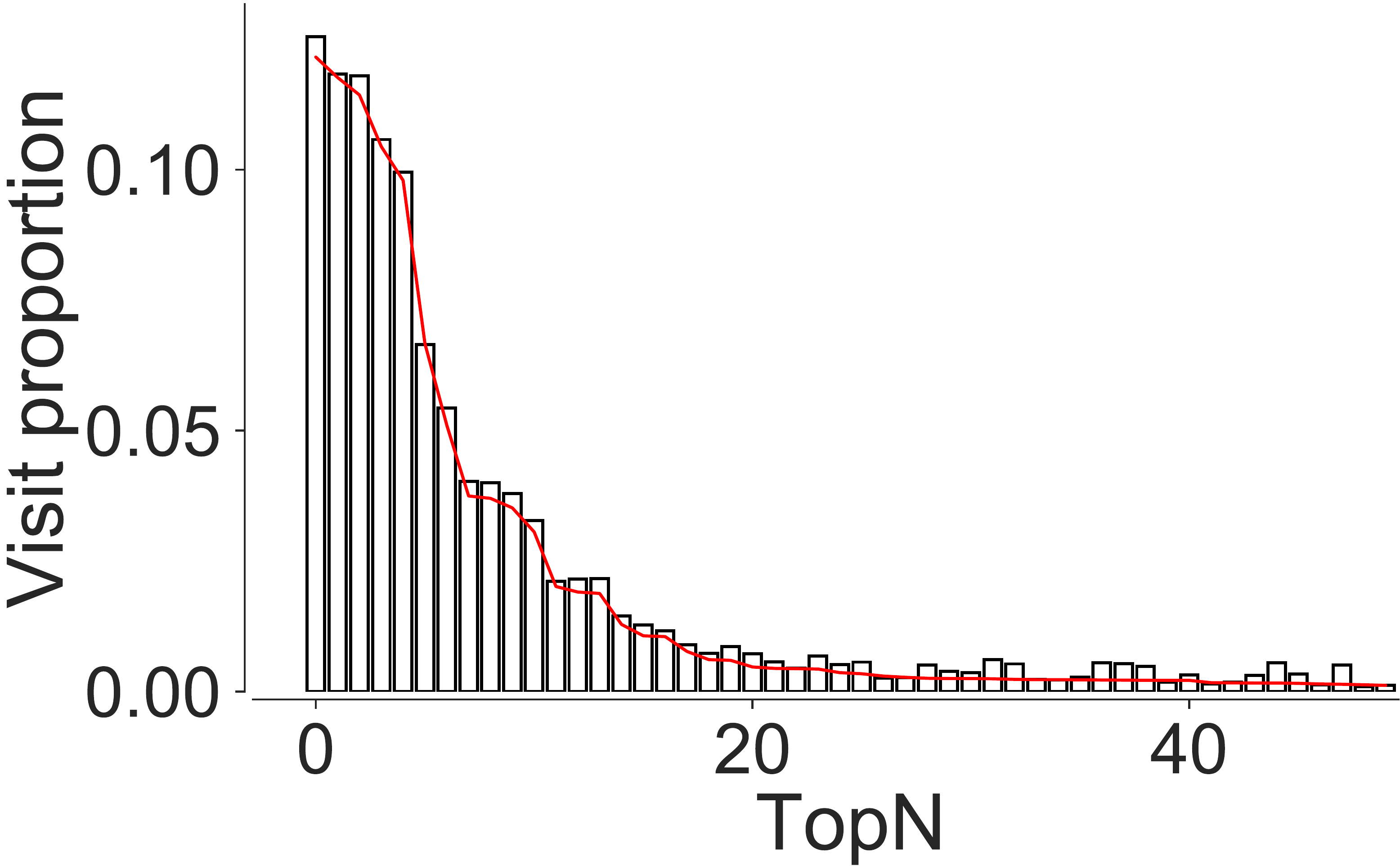} }}%
    \subfloat[Copula]{{\includegraphics[scale=0.1]{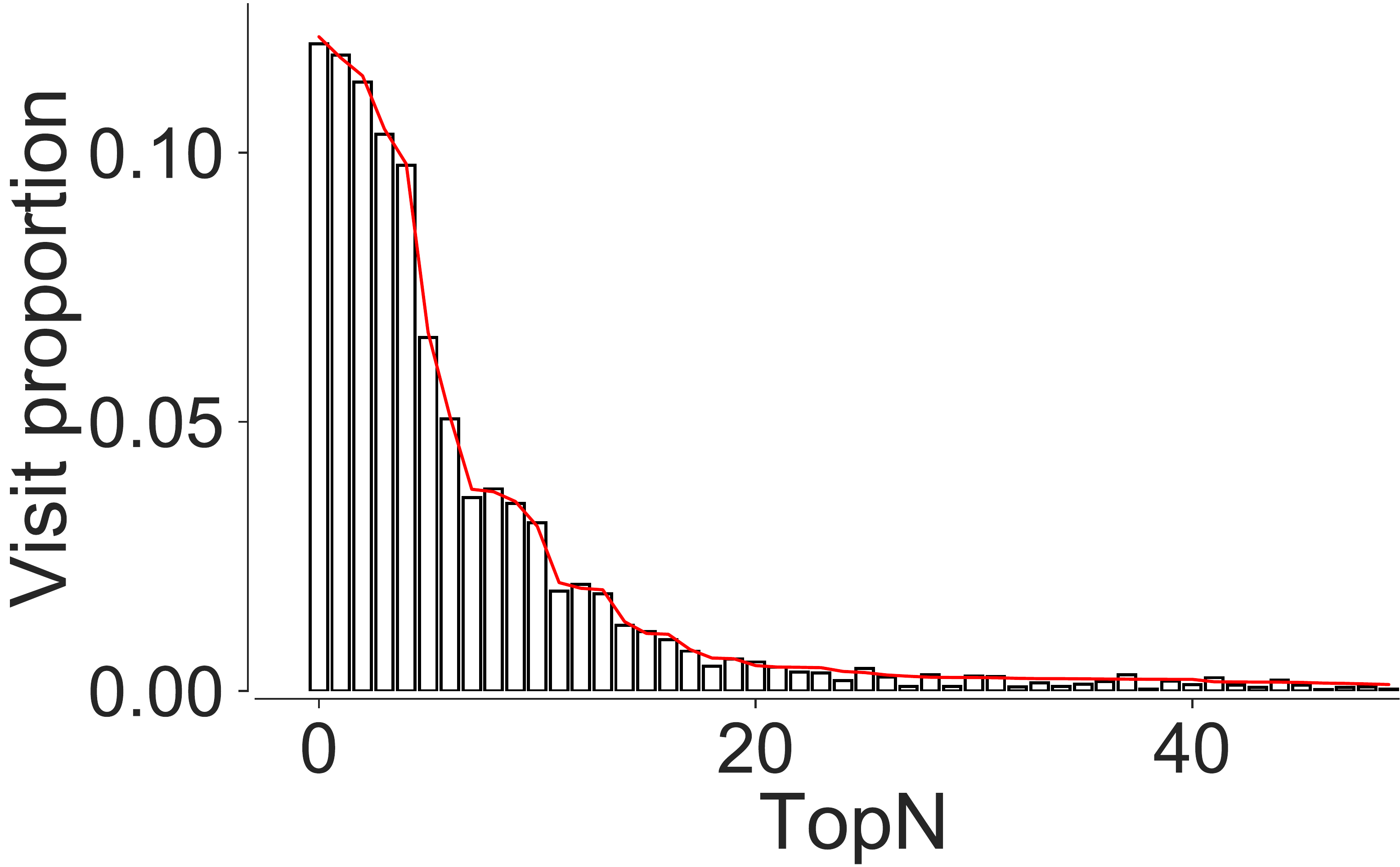} }}}
    \newline
    \caption{TopN visited locations for real and synthetic trajectories generated by each method. We select $N=50$ out of a total of 286 locations. The red curve shows the distribution for the true dataset.}%
    \label{fig:geo_sem}%
\end{figure}

\textbf{Experimental setup} 
A complete trajectory sequence can be generated by iteratively feeding the current output trajectory sequence as input for the next step to the trained model.
RNNs are trained on the geo-hashes and timestamps of all the individuals present in the dataset in a deterministic framework.
GANs are first trained to model and then successively reproduce the traces in the same representation, which is mapped back to the $(lat_i,lon_i)$ coordinates.
We use the standard implementations of the predictive algorithms and hyper-parameters as described in their respective papers.
To use copulas as generative models, we rely on the \emph{rvinecopulib} package~\citep{rvinecopulib}, whose \texttt{vine} routine implements the automatic kernel-based fitting of the dependence structure.

\textbf{Dataset} 
Experiments are performed using the Nokia mobile dataset~\cite{laurila2012mobile} that consists of mobility trajectories of individuals collected in Switzerland. 
We use a total of 70M data points to train the considered models.

\textbf{Evaluation} 
We perform the evaluation of the generated trajectories using this dataset from four distinct dimensions: (1) geographic and semantic similarity, (2) statistical similarity (3) long-range dependencies and (4) privacy tests.
In order to assess the geographic and semantic similarity, we compare the probability distribution of visiting $topN$ locations (visit-time and dwell-time) in the generated trajectories for each technique compared to the true dataset (see~\autoref{fig:geo_sem}). 
Char-RNN, RGAN and copulas have the closest fit to the true distribution indicating that the $topN$ locations are very well preserved in the respective synthetic datasets.

\begin{figure}[!t]
\resizebox{\textwidth}{!}{
    \centering
    \subfloat[Char-RNN]{{\includegraphics[scale=0.1]{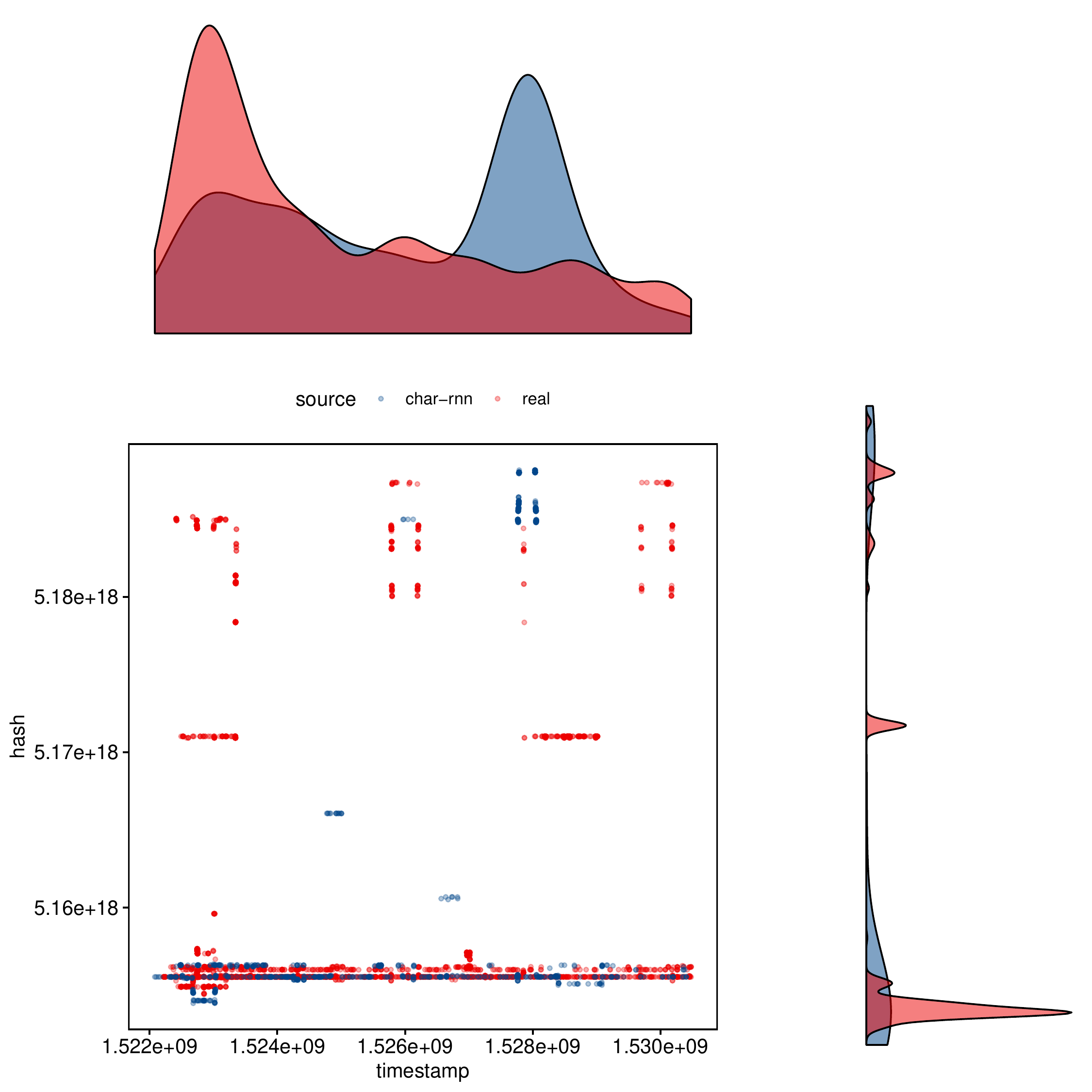} }}%
    \subfloat[RNN-LSTM]{{\includegraphics[scale=0.1]{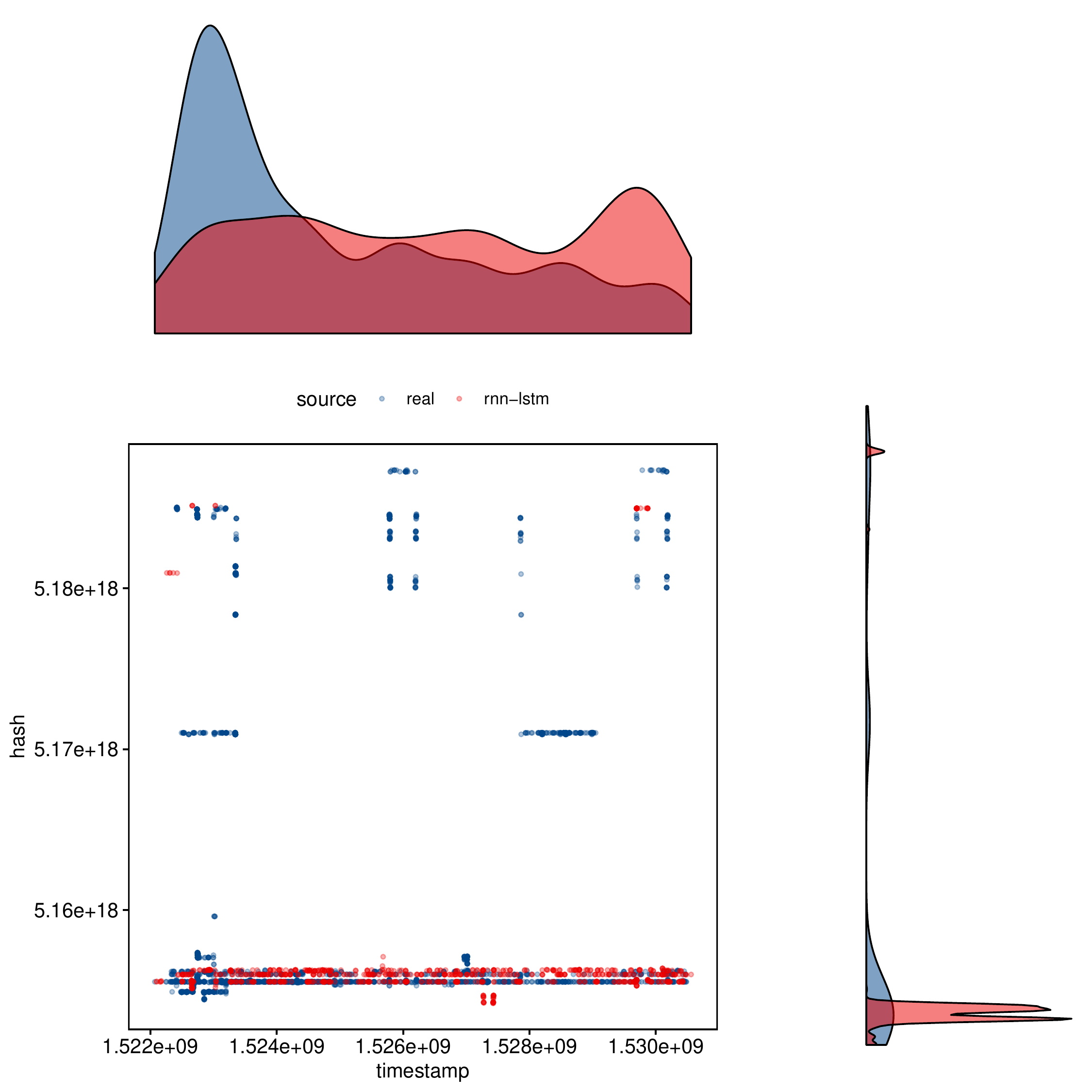} }}%
    \subfloat[RHN]{{\includegraphics[scale=0.1]{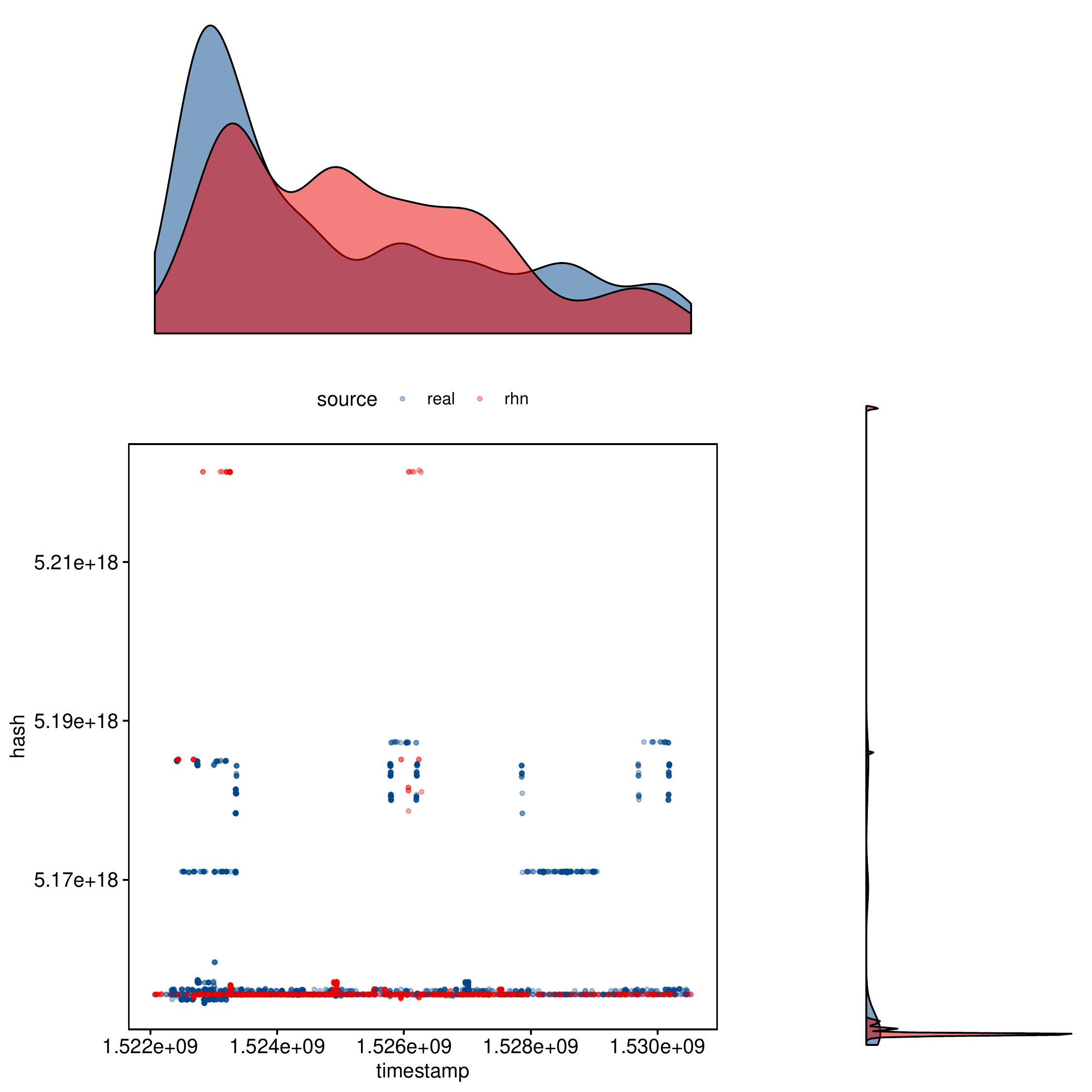} }}\newline
    \subfloat[PSMM]{{\includegraphics[scale=0.1]{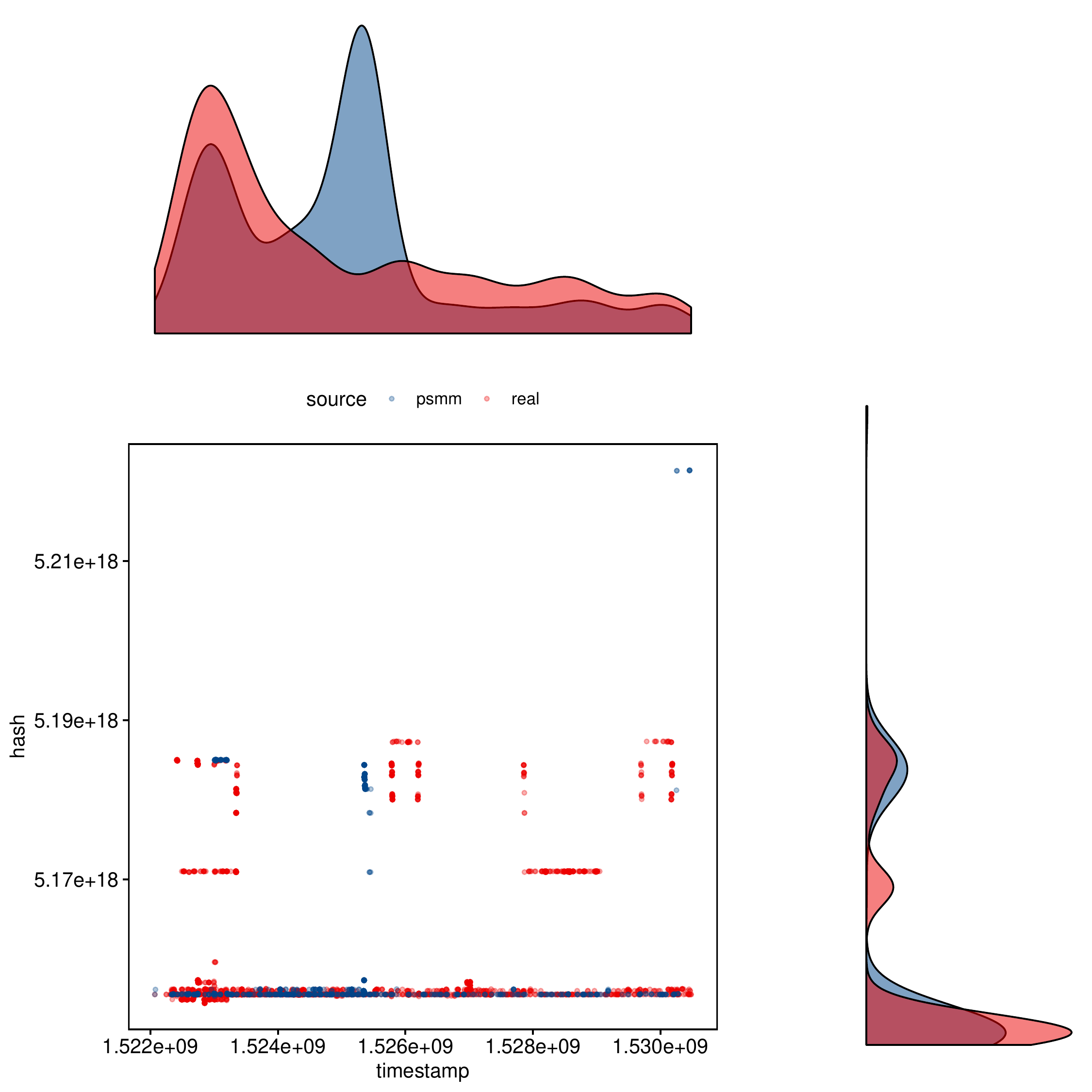} }}
    \subfloat[SGAN]{{\includegraphics[scale=0.1]{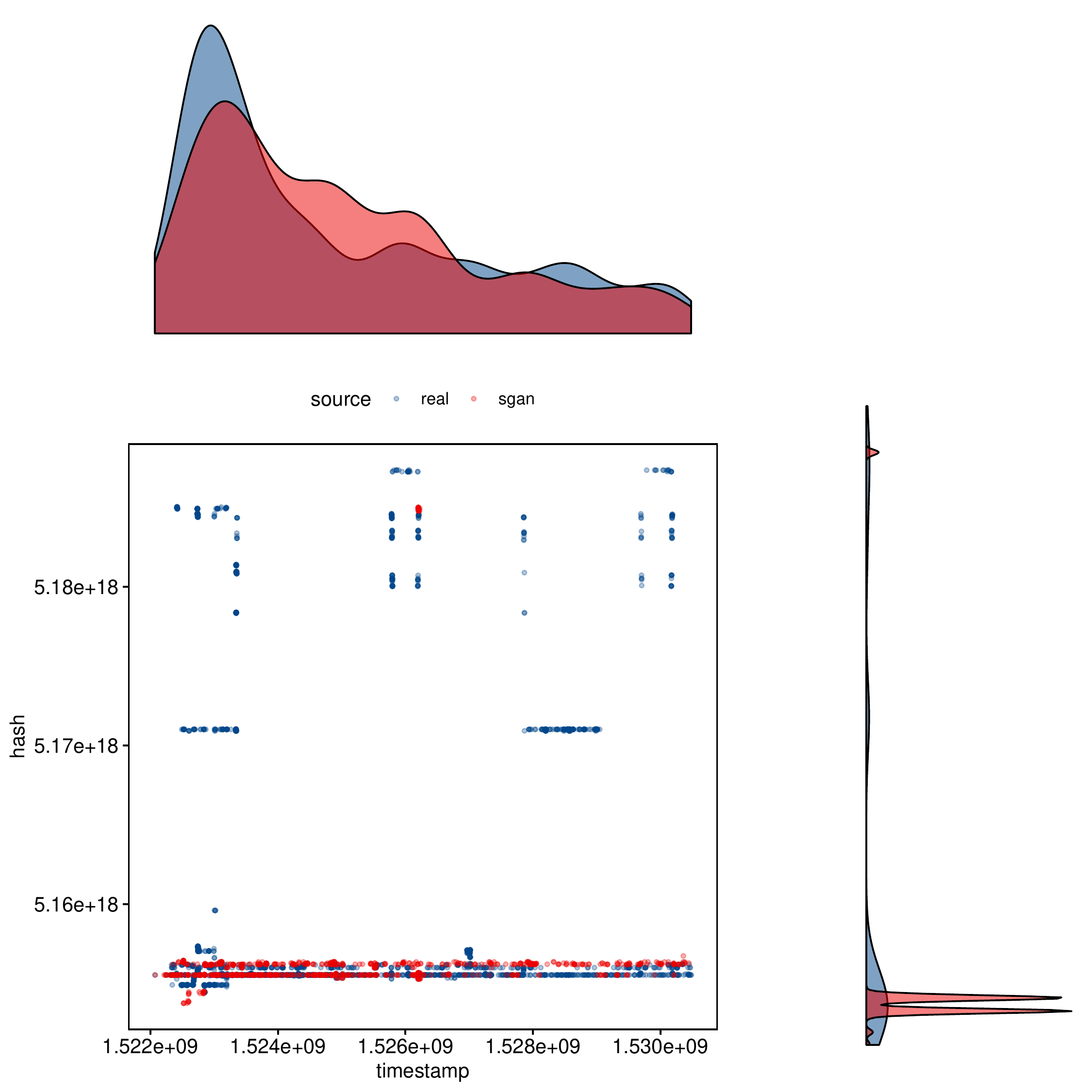} }}%
    \subfloat[RGAN]{{\includegraphics[scale=0.1]{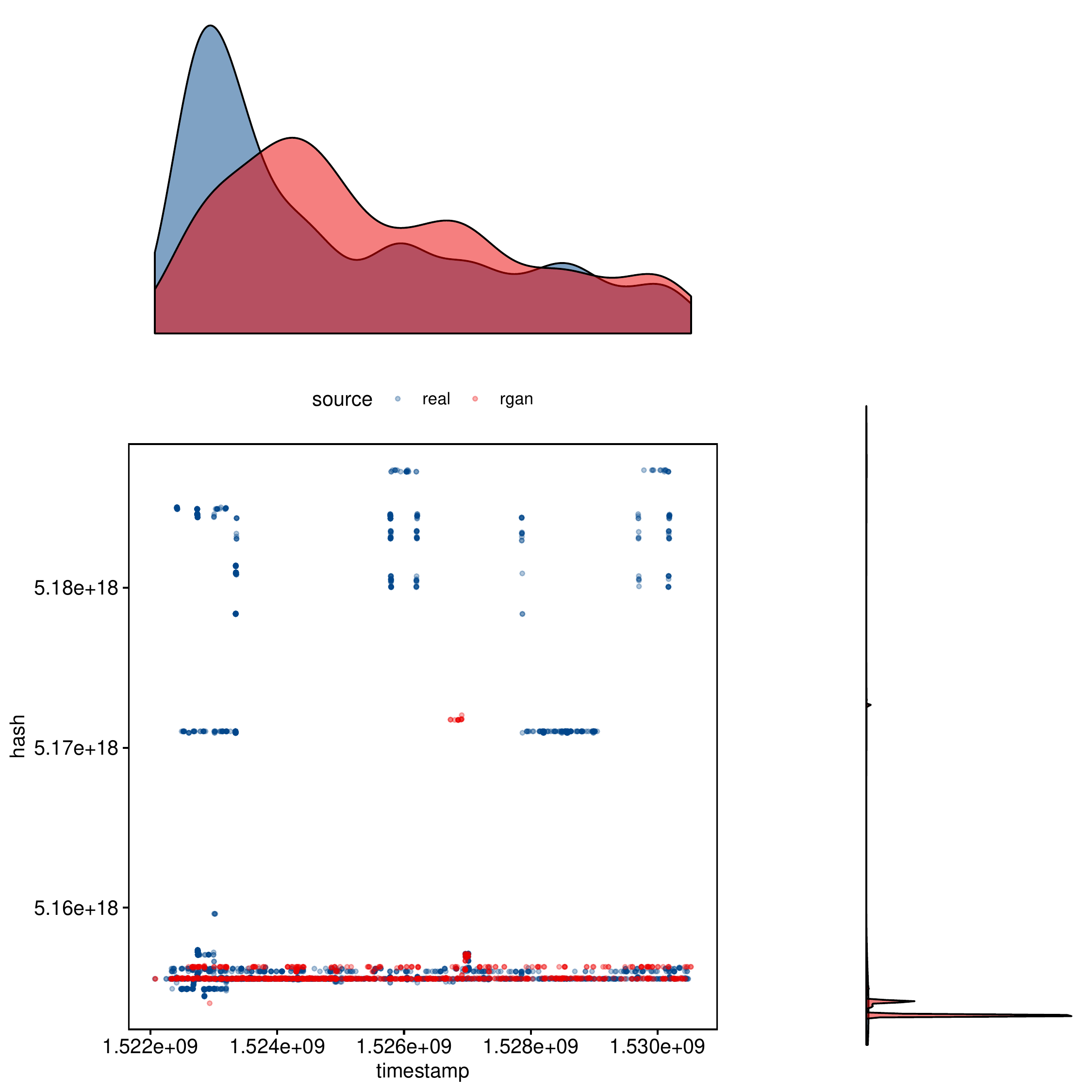}  }}%
    \subfloat[Copula]{{\includegraphics[scale=0.1]{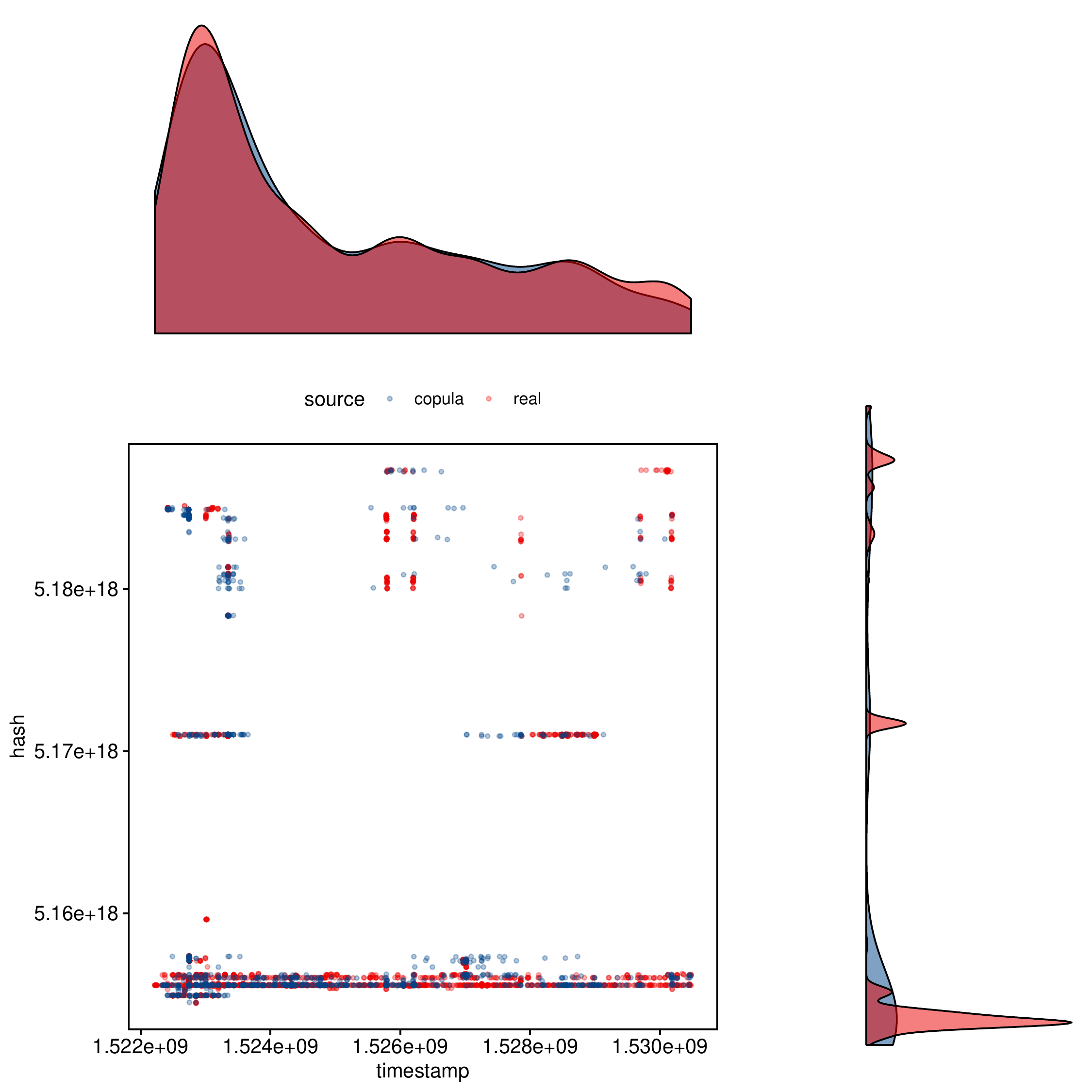} }}}
    \newline
    \caption{Scatter plots}%
    \label{fig:scatter}%
\end{figure}

To evaluate the statistical similarity, we use Mean Maximum Discrepancy (MMD)~\citep{gretton2012kernel} to test whether one can reject the null hypothesis that a synthetic sample has the same distribution as the data.
The scatter plots are shown in~\autoref{fig:scatter}.
MMD works by replacing the probability densities with embeddings that facilitate the computation of distances between distributions.
Note that defining distance metrics in the context of time series data such as mobility trajectories is challenging due to the alignment concerns~\citep{esteban2017real}.
We thus consider the time axis for alignment as done by~\citet{esteban2017real}.
The results along with the training and generation time for each approach is shown in~\autoref{tab:two-sample-metrics}, where we observe that all approaches achieve similar results in terms of MMD, with copulas standing out with a lower value.
We can thus infer that copulas can synthesize distributions with statistical characteristics closer to the observed ones.
Regarding the computational efficiency, copulas require a fraction of the time needed by NN-based approaches.

\begin{table}[h!]
\centering
\caption{Mean and standard deviation of real vs. synthetic data (lower is better) from 30 repetitions. Second row is CPU time indicating the training/fit+generation time.}
\resizebox{\textwidth}{!}{
\begin{tabular}{@{}lccccccc@{}}
\toprule
\textbf{Metric/Method} & \multicolumn{1}{c}{\textbf{Char-RNN}} & \multicolumn{1}{c}{\textbf{RNN-LSTM}} & \multicolumn{1}{c}{\textbf{RHN}} & \multicolumn{1}{c}{\textbf{PSMM}} & \multicolumn{1}{c}{\textbf{SGAN}} & \multicolumn{1}{c}{\textbf{RGAN}} & \multicolumn{1}{c}{\textbf{Copula}} \\ \midrule
\textbf{MMD}           & \multicolumn{1}{c}{0.32(1e-3)}        & \multicolumn{1}{c}{0.27(9e-4))}        & \multicolumn{1}{c}{0.30(1e-3)}   & \multicolumn{1}{c}{0.21(6e-4)}    & \multicolumn{1}{c}{0.19(7e-4)}    & \multicolumn{1}{c}{0.21(6e-4)}    & \multicolumn{1}{c}{0.01(6e-4)}       \\
\textbf{CPU time (sec)}      & 9k+$\sim$10                                       & 10.3k+$\sim$14                                       & 12.7k+$\sim$15                                & 10.5k+$\sim$15                                & 11.2k+$\sim$15                                 & 11.5k+$\sim$14                                & 6.5 + 0.76                       \\ \bottomrule
\end{tabular}}
\label{tab:two-sample-metrics}
\end{table}

\begin{figure}[!t]
    \centering
    \subfloat[Mutual information decay]{{\includegraphics[scale=0.15]{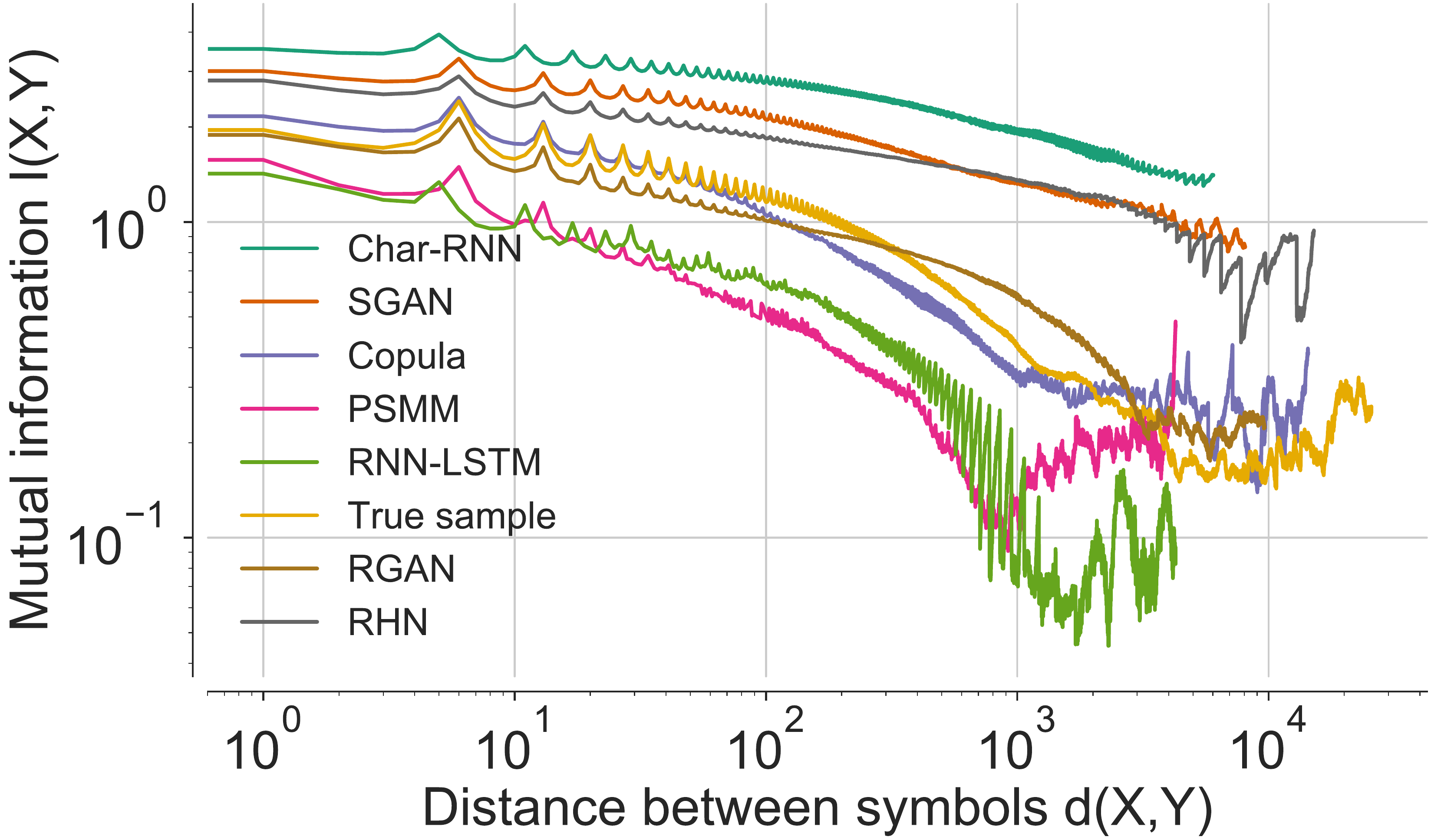} }}%
    \subfloat[Privacy tests]{{\includegraphics[scale=0.14]{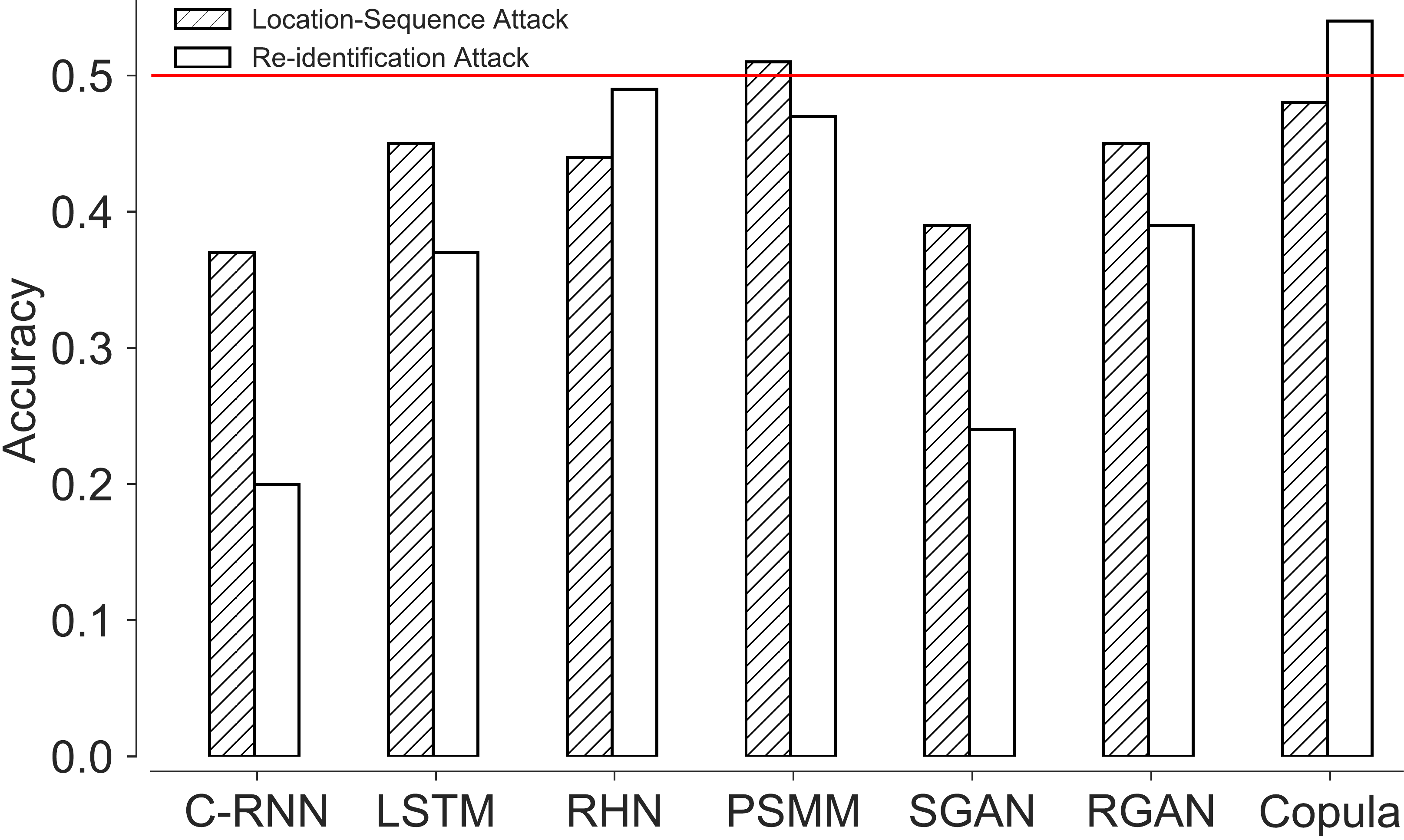} }}
    \subfloat[Sample trajectories]{{\includegraphics[scale=0.085]{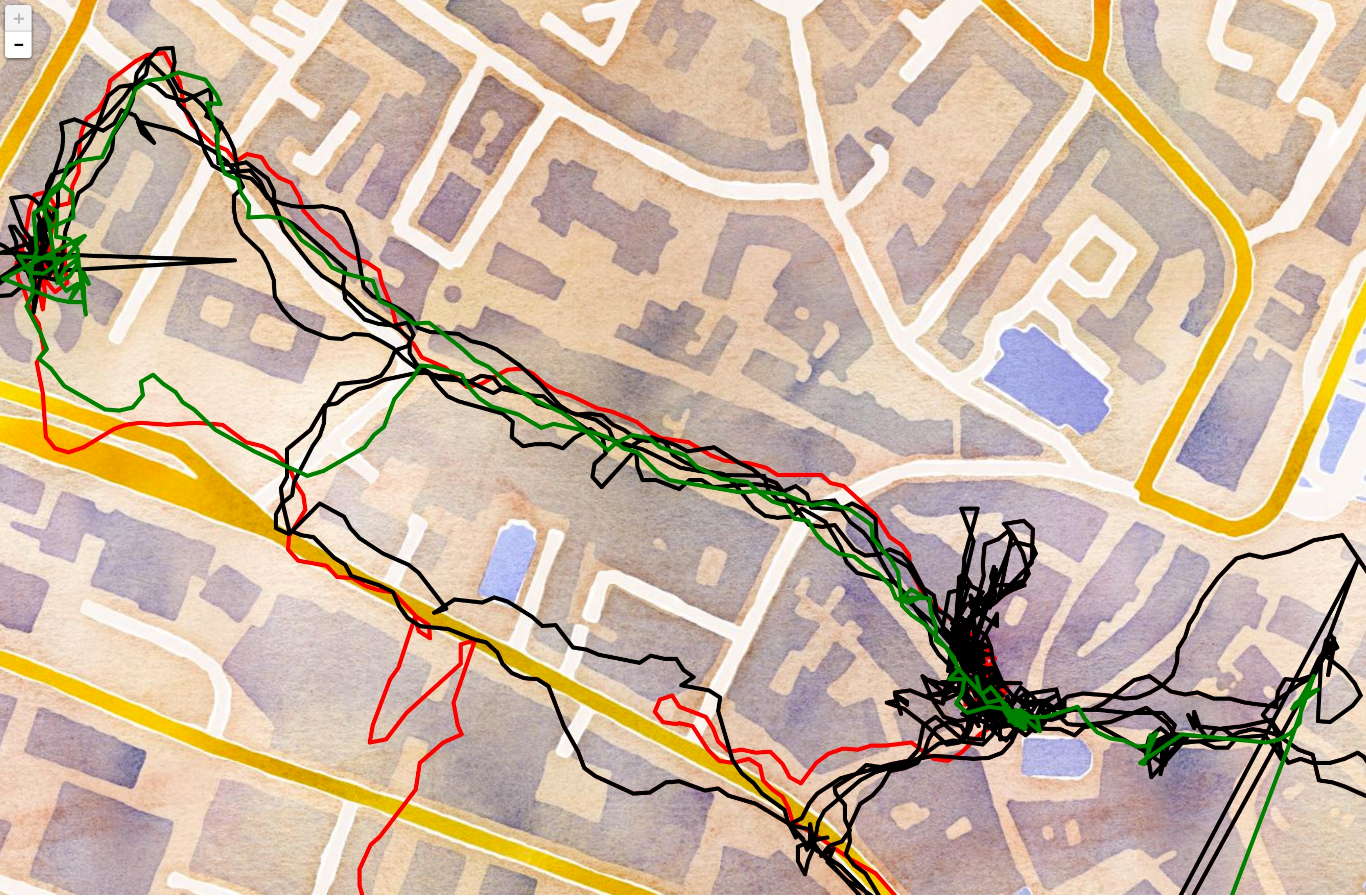} }}
    \caption{(a) long-range dependency test (symbols denote individual location coordinates), (b) privacy test with location hiding as privacy preserving mechanism (red line indicates a random guess) (c) sample trajectories generated by two best approaches (copulas (black) and SGANs (red)) follow the road network for the most part and also synthesize stays at some locations indicating a point of interest. Trajectory from the actual dataset in the same area is depicted in green.}    
    \label{fig:pri_mi}%
\end{figure}

Figure~\ref{fig:pri_mi}(a) shows the result of long-range dependency test, in terms of mutual information decay~\cite{mahalunkar2018understanding, lin2016critical}.
We observe a power-law decay in case of GANs, copulas and RNN-LSTM indicating that they account for the long-range dependencies in mobility trajectories.
Figure~\ref{fig:pri_mi}(b) shows the results of two privacy tests: (1) location-sequence attack, and (2) membership interference attack.
Given a synthetic dataset, (1) answers to what level of accuracy can trajectories in the dataset be reconstructed~\cite{shokri2011quantifying}, and (2) an adversary's accuracy of inferring if a target individual contributed to the specific trajectory~\cite{pyrgelis2017knock}.  
For these tests, we use the the location-privacy and mobility meter~\cite{shokri2011quantifying}, where obfuscation is performed using the location hiding mechanism.
Given a completely random distribution the accuracy of a recovered user-information is 0, we therefore suspect that the privacy-based score is biased towards representations which do not accurately capture the statistical properties of the true dataset.

\section{Conclusion and Future Work}

In this work, we propose and evaluate a variety of generative models to synthesize mobility trajectories. 
To the best of our knowledge, this is the first study to do so using seven different approaches while evaluating their realism across four dimensions.  
From the results and discussion, we observe that regarding statistical and semantic properties, copulas have an advantage over all other methods. 
Additionally, all NN-based methods are time consuming, which makes copulas favorable when computational efficiency is important to the end-users.
As future work, we will consider datasets collected in bigger cities and generate larger synthetic datasets to evaluate the performance of these models under high movement stochasticity.
From curbing the privacy leakage of the true dataset while maintaining utility, trajectory generation can be designed as an optimization problem with an objective to jointly maximize statistical similarity and privacy.
But it is still not clear how to assess such property and adaptive/configurable metric, and is part of ongoing work~\cite{nasr2018machine}.
While this paper represents an initial comparative study of various generative models, a deeper understanding of their performances will be needed to compute utility-privacy scores as applied to online services by~\citet{krause2008utility} before publicly releasing the synthetic datasets.
Another interesting avenue for research is to apply transfer learning in order to map a mobility behavioral model captured in one city on to another region.

\bibliographystyle{plainnat}
\bibliography{nips18}

\end{document}